\documentclass{ieeeaccess}
\usepackage{cite}
\usepackage{amsmath,amssymb,amsfonts}
\usepackage{algorithmic}
\usepackage{graphicx}
\usepackage{textcomp}
\usepackage{tabularray}
\usepackage{booktabs}
\usepackage{placeins}
\usepackage{float}

\usepackage{bm}
\usepackage{hyperref}

\makeatletter
\AtBeginDocument{\DeclareMathVersion{bold}
\SetSymbolFont{operators}{bold}{T1}{times}{b}{n}
\SetSymbolFont{NewLetters}{bold}{T1}{times}{b}{it}
\SetMathAlphabet{\mathrm}{bold}{T1}{times}{b}{n}
\SetMathAlphabet{\mathit}{bold}{T1}{times}{b}{it}
\SetMathAlphabet{\mathbf}{bold}{T1}{times}{b}{n}
\SetMathAlphabet{\mathtt}{bold}{OT1}{pcr}{b}{n}
\SetSymbolFont{symbols}{bold}{OMS}{cmsy}{b}{n}
\renewcommand\boldmath{\@nomath\boldmath\mathversion{bold}}}
\makeatother

\def\BibTeX{{\rm B\kern-.05em{\sc i\kern-.025em b}\kern-.08em
    T\kern-.1667em\lower.7ex\hbox{E}\kern-.125emX}}

\begin{document}

\title{Applying LLMs to Active Learning: Towards Cost-Efficient Cross-Task Text Classification without Manually Labeled Data}

\author{\uppercase{YEJIAN ZHANG}\authorrefmark{1}, 
\uppercase{SHINGO TAKADA}\authorrefmark{2}}

\address[1]{Graduate School of Science and Technology, Keio University, Yokohama, Japan (e-mail: zhangyejian@doi.ics.keio.ac.jp)}
\address[2]{Graduate School of Science and Technology, Keio University, Yokohama, Japan (e-mail: michigan@ics.keio.ac.jp)}


\corresp{Corresponding author: YEJIAN ZHANG (e-mail: zhangyejian@doi.ics.keio.ac.jp).
}

\begin{abstract}

Machine learning-based classifiers have been used for text classification, such as sentiment analysis, news classification, and toxic comment classification. However, supervised machine learning models often require large amounts of labeled data for training, and manual annotation is both labor-intensive and requires domain-specific knowledge, leading to relatively high annotation costs. To address this issue, we propose an approach that integrates large language models (LLMs) into an active learning framework, achieving high cross-task text classification performance without the need for any manually labeled data. Furthermore, compared to directly applying GPT for classification tasks, our approach retains over 93\% of its classification performance while requiring only approximately 6\% of the computational time and monetary cost, effectively balancing performance and resource efficiency. These findings provide new insights into the efficient utilization of LLMs and active learning algorithms in text classification tasks, paving the way for their broader application.

\end{abstract}


\titlepgskip=-21pt

\maketitle


\section{Introduction}
Text classification is a fundamental task in natural language processing (NLP), involving the automatic categorization of textual data into predefined categories. It plays a critical role in applications such as sentiment analysis, news classification, and toxic comments detection. These applications span multiple domains and are essential for efficiently processing and analyzing large volumes of text, enabling organizations to extract insights from data. As the volume of textual information grows, the importance of accurate text classification methods has increased rapidly, making it a vital component of NLP research and industry applications \cite{survey2022, deepLearningTC}.

Machine learning-based supervised text classifiers have been widely used and have proven effective in performing text classification tasks. However, these models typically require large amounts of labeled data for training, which presents significant challenges. Manual annotation of labeled data is both labor-intensive and knowledge-intensive, as it requires annotators to have domain-specific expertise to ensure accurate labeling~\cite{settles.tr09}. This process is time-consuming and costly~\cite{ratner2017snorkel,tomanek2009semi}, which makes it difficult to scale the development of high-quality training datasets.

The development of Large Language Models (LLMs) has introduced models like the Generative Pre-trained Transformer (GPT), which are capable of performing zero-shot classification tasks through a question-and-answer format. However, these models were not specifically designed for large-scale text classification tasks, which leads to several limitations in such applications \cite{strubell2019energy, bender2021dangers}. As a large-scale generative model, GPT requires substantial computational resources for each task execution, particularly because classification predictions are made by generating text sequentially through a question-and-answer process \cite{brown2020language}. Moreover, inefficiencies in batch processing, delays in model invocation and response times, and the current processing rate limitations of advanced GPT models further exacerbate the bottlenecks when directly applying GPT to large-scale classification tasks, resulting in higher costs and longer run times \cite{Narayanan2021, Rae2021,10433480}.

To address the aforementioned limitations, we propose an LLM-driven active learning framework. The core idea is to employ a generative LLM (GPT), guided by structured prompts, to annotate instances selected by active learning query strategies, replacing the traditional human oracle. In addition, we evaluate several distinct query strategies and leverage RoBERTa to generate text embeddings that support the query selection process and the training of downstream machine learning classifiers. This design enables our approach to perform cross-task text classification tasks without any human-annotated data, while avoiding the disadvantages associated with using GPT for large-scale text classification tasks directly. To validate our approach, we have tested our approach on several datasets including IMDB dataset \cite{maas-EtAl:2011:ACL-HLT2011} for sentiment classification, AGnews dataset \cite{zhang2015character} for news categorization, and Jigsaw Toxic Comment Classification dataset \cite{jigsaw2018toxic} for toxic comment classification.

The main contributions of this work are as follows:
\begin{enumerate}
    \item We propose an LLM-driven active learning approach for cross-task text classification without relying on human-annotated data.
    \item By incorporating GPT into an active learning framework, we mitigate the limitations associated with directly using GPT for large-scale classification tasks, balancing performance and resource efficiency.
    \item We demonstrate the effectiveness of our approach through extensive experiments on multiple tasks, including sentiment classification, news classification, and toxic comment classification, achieving high performance across tasks.
\end{enumerate}

\section{Related Work}
\subsection{Importance of Text Classification}

Text classification has been widely studied for its critical role in natural language processing. Sebastiani \cite{sebastiani2002machine} provided a comprehensive survey of text categorization techniques, highlighting its importance in organizing and managing the growing amount of unstructured text data across various domains. Pang et al.  \cite{pang2002thumbs} demonstrated the significance of text classification in sentiment analysis, which helps to understand public opinion by automatically determining the sentiment conveyed in reviews or social media posts. Androutsopoulos et al. \cite{androutsopoulos2000an} applied text classification to spam detection, showing its effectiveness in filtering unwanted messages. Similarly, Lewis and Gale \cite{lewis1994sequential} investigated topic categorization methods for efficient information retrieval, which are critical in enabling better decision-making in fields such as healthcare, finance, and media. These studies demonstrate the role of text classification in automating processes, extracting insights, and managing information efficiently.

\subsection{Machine Learning-Based Text Classification}

Machine learning methods have been widely applied to text classification. Sebastiani \cite{sebastiani2002machine} provided an early comprehensive survey on automated text categorization using machine learning, detailing the use of algorithms like Naive Bayes and support vector machines, and emphasizing the importance of feature extraction techniques such as TF-IDF. Kowsari et al. \cite{kowsari2019text} reviewed traditional machine learning approaches as well as recent deep learning advancements in text classification, highlighting the shift from manual feature engineering to automatic representation learning. Similarly, Zhang and Wallace \cite{zhang2015text} conducted a survey of text classification techniques, discussing both conventional models and emerging deep learning frameworks, while pointing out the challenges related to labeled data and model scalability. These reviews illustrate the progression of machine learning methods in text classification, while acknowledging the dependency on large labeled datasets.

\subsection{Large Language Models}
\subsubsection{Bidirectional Transformer-based Models}
Devlin et al. \cite{devlin2018bert} introduced BERT (Bidirectional Encoder Representations from Transformers), which employs a bidirectional attention mechanism to capture contextual information from both preceding and following tokens. This bidirectional modeling significantly enhances the quality of learned text embeddings, making them more suitable for a wide range of NLP tasks. 

Liu et al. \cite{liu2019roberta} proposed RoBERTa, an optimized variant of BERT, which eliminates the Next Sentence Prediction (NSP) objective and incorporates dynamic masking and larger batch sizes to improve pretraining efficiency. These modifications allow RoBERTa to learn more robust and contextually rich text representations, leading to improved performance in downstream applications. Specifically, RoBERTa has been shown to generate high-quality text embeddings, making it a strong choice for tasks that rely on effective feature extraction rather than generative text modeling.

\subsubsection{Generative Pre-trained Transformer}
Brown et al. \cite{brown2020language} introduced GPT-3, a state-of-the-art language model capable of performing various NLP tasks such as text completion, translation, and summarization, with minimal or no task-specific training data. This model exemplifies zero-shot learning capabilities, where extensive pretraining on diverse datasets allows it to generalize across different tasks by leveraging its broad knowledge base rather than explicit supervision.

While generative AI models like GPT have demonstrated impressive performance across multiple domains, their application in structured classification tasks at scale faces certain limitations and challenges.
Rae et al. \cite{Rae2021} analyzed the computational trade-offs of scaling large language models, highlighting that while larger models achieve improved performance, they also introduce inefficiencies in inference and deployment. The study shows that GPT-style models, due to their autoregressive nature, require sequential token generation, limiting parallelization and increasing inference latency.
Moreover, Raiaan et al. \cite{10433480} provided a comprehensive analysis of large language models (LLMs), highlighting both their advancements and inherent challenges. Their study emphasizes that while models like GPT demonstrate strong performance across various NLP tasks, their deployment at scale faces computational and efficiency constraints. In addition to discussing the limitations posed by the autoregressive decoding mechanism, they further elaborate on the challenges introduced by high computational cost and energy consumption in real-world applications.

\subsection{Active Learning}

Active learning has been widely studied as an effective approach to reduce labeling costs by selecting the most informative samples for annotation. Angluin \cite{angluin1988queries} introduced query-based learning in a theoretical framework. Seung et al. \cite{QBC} proposed Query by Committee (QBC), where a set of models (a “committee”) is trained on the same labeled data, and the instances that induce the greatest disagreement among them are selected for labeling. Cohn et al. \cite{cohn1994improving} later formalized active learning as a general framework, proposing a query-based selection approach to improve model generalization with fewer labeled examples.

Various query strategies have been introduced to improve active learning effectiveness. Brinker \cite{10.5555/3041838.3041846} explored the integration of diversity-aware selection into uncertainty-based active learning, ensuring that queried samples are both uncertain and well-distributed in the feature space, thereby reducing redundancy in selected data. Sener and Savarese \cite{sener2018active} developed core-set sampling, a geometric approach that formulates active learning as a coverage problem, selecting samples that best represent the entire dataset. Another approach, information density sampling, was introduced by Settles and Craven \cite{settles-craven-2008-analysis}, where samples are selected based on both uncertainty and density in the feature space, prioritizing representative and high-impact instances.

Active learning has been applied across various domains to enhance model performance while minimizing labeling efforts. Tong and Koller \cite{tong2002support} explored the integration of active learning with support vector machines in text classification, demonstrating that selective sampling significantly reduces the number of labeled instances required for accurate classification. 

Joshi et al.\cite{5206627} introduced an active learning approach for multi-class image classification, demonstrating that balancing uncertainty and representativeness improves labeling efficiency in visual recognition tasks. In the realm of structured prediction tasks, Settles and Craven \cite{settles-craven-2008-analysis} analyzed various active learning strategies for sequence labeling, such as named entity recognition and part-of-speech tagging, highlighting the benefits of selecting informative sequences to improve model accuracy.

\section{Proposed Approach}
\label{sec:III}
\begin{figure*}[]
  \centering
  \includegraphics[width=0.9\linewidth]{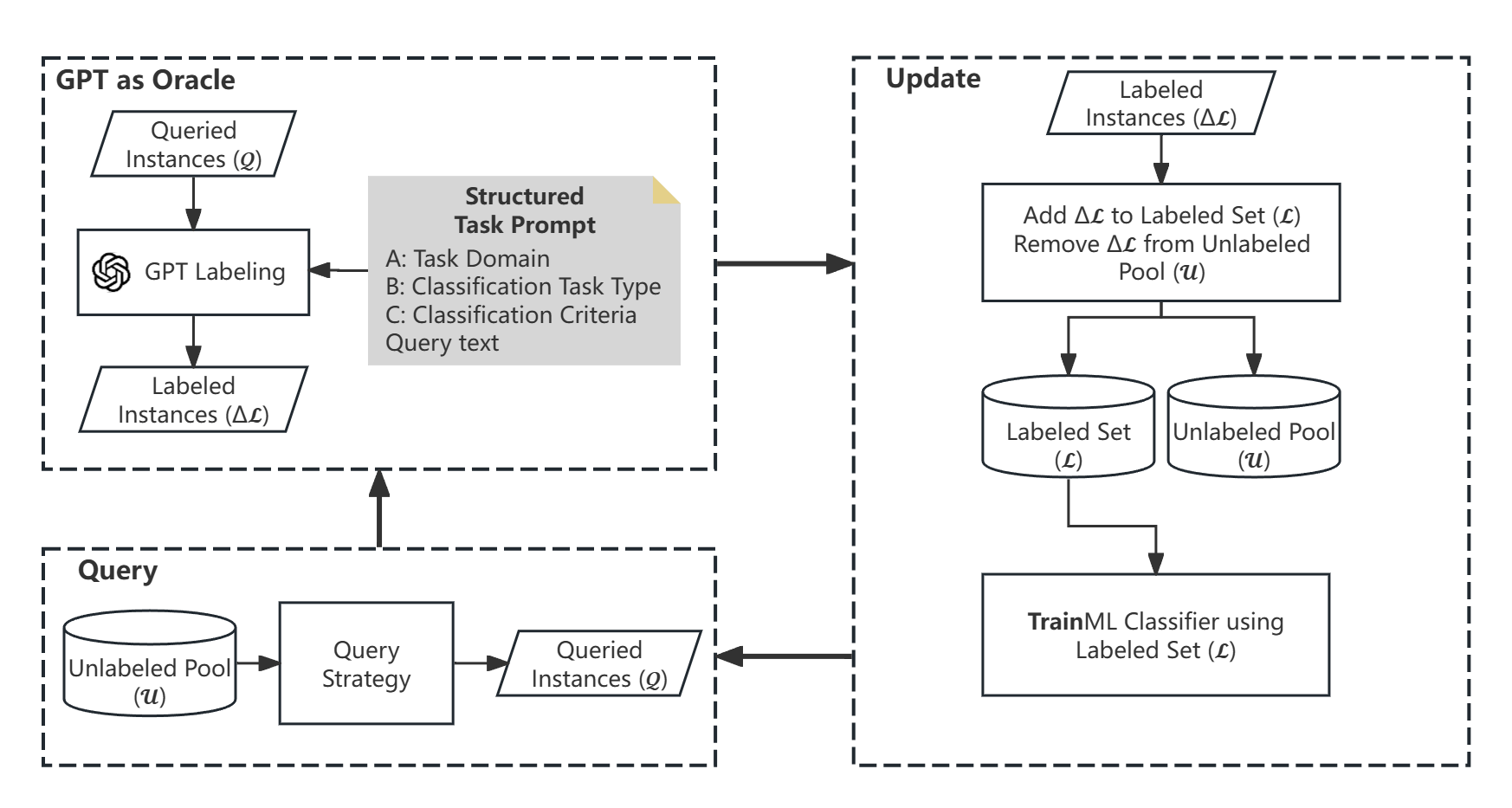}
  \caption{Overall Workflow: LLM-driven Active Learning Loop}
  \label{Fig1}
\end{figure*}

We propose an LLM-driven active learning framework designed to perform cross-task text classification without relying on human-annotated data. The overall workflow of our approach is illustrated in Figure~\ref{Fig1}. Within the active learning loop, a query strategy selects informative instances from the unlabeled pool \( \mathcal{U} \) and sends them to the GPT module for annotation. Guided by structured prompts, the GPT module labels the queried instances \( Q \). The resulting labeled instances \( \Delta\mathcal{L} \) are then added to the labeled set \( \mathcal{L} \), which is used to update the machine learning classifier. In the following sections, we detail each component of this loop.

\subsection{Query and Update}

In the Query phase, a query strategy selects a batch of the most informative instances from the current unlabeled pool \( \mathcal{U} \), which are then passed to the oracle for labeling.

In the Update phase, the newly labeled instances \( \Delta \mathcal{L}\) are added to the labeled set \( \mathcal{L}\) and removed from the unlabeled pool \( \mathcal{U}\). The machine learning classifier is retrained on the updated labeled set. Before the first iteration, a seed set is randomly selected from unlabeled pool \( \mathcal{U}\) and labeled by the oracle to initialize the query strategy.

We implemented four widely studied query strategies: uncertainty sampling (Query by Committee)~\cite{QBC}, entropy sampling with diversity~\cite{lewis1994sequential, settles.tr09}, core-set sampling~\cite{sener2018active}, and information density sampling~\cite{settles-craven-2008-analysis}. Each strategy is evaluated independently in our experiments. We set the logistic regression classifier (\(solver=lbfgs\), \(multi\_class=auto\), \(max\_iter=1000\)) as the default machine learning classifier, except for QBC, which uses a voting classifier to form the committee. Additionally, to support query selection and classifier training, we utilize the roberta-base model with mean pooling to generate fixed-dimensional contextual embeddings. The subsequent subsections provide detailed descriptions of each query strategy.

\subsubsection{Uncertainty Sampling (Query by Committee)}
For uncertainty sampling, we implemented a Query by Committee (QBC) approach using a voting classifier. The committee consists of four classifiers, including an SVM (\(Kernel=linear\), \(Probability=True\)), a decision tree (\(Max Depth=None\)), a random forest (\(Estimators=100\), \(Max Features=None\)), and a logistic regression (\(solver=lbfgs\), \(multi\_class=auto\), \(max\_iter=1000\)) model. We constructed a soft voting classifier using these four models to better reflect the overall uncertainty, with the calculation of voting entropy given in Equation (\ref{Equ1}):
\begin{equation}
\label{Equ1}
H(x_i) = - \sum_{j=1}^{C} \left( \frac{1}{M} \sum_{m=1}^{M} p_{ij}^{(m)} \right) \log \left( \frac{1}{M} \sum_{m=1}^{M} p_{ij}^{(m)} \right)
\end{equation}
where \( H(x_i) \) is the entropy of sample \( x_i \), \( C \) is the total number of classes, \( M \) is the number of models in the committee, and \( p_{ij}^{(m)} \) is the probability predicted by model \( m \) that sample \( x_i \) belongs to class \( j \). The samples with the highest voting entropy are selected, representing the instances where the committee models are most uncertain.

\subsubsection{Entropy Sampling with Diversity}
For entropy sampling with diversity, we use the logistic regression model to predict the probabilities of the unlabeled samples. The entropy for each sample is computed based on the predicted probabilities in equation (\ref{Equ2})
\begin{equation}
\label{Equ2}
H(x_i) = - \sum_{j=1}^{C} p_{ij} \log p_{ij}
\end{equation}
where \( H(x_i) \) represents the uncertainty of sample \( x_i \), \( C \) is the total number of classes, and \( p_{ij} \) is the probability predicted by the Logistic Regression model that sample \( x_i \) belongs to class \( j \). The samples with the highest entropy are initially selected, representing those with the greatest uncertainty. To ensure diversity among the selected samples, pairwise distances are computed between the high-entropy candidates using RoBERTa embeddings, and diversity scores are assigned based on these distances. The final selection is determined by combining entropy scores and diversity scores for each sample, where the entropy score and diversity score are summed to produce a combined score. The samples with the highest combined scores are selected, resulting in a set that is both uncertain and diverse, thereby improving the model's generalizability by covering a broader region of the input space.

\vspace{-1em}
\subsubsection{Core-set sampling}
For core-set sampling, we use the pre-computed embeddings from the RoBERTa embedder to map samples into a high-dimensional feature space. Pairwise distances are then computed between the unlabeled samples and the already labeled samples to determine the minimum distance for each unlabeled sample. The goal is to select samples that are farthest from the labeled set, thereby ensuring coverage of different regions of the feature space. Specifically, the samples with the largest minimum distance are chosen, allowing the model to expand its understanding by exploring less-represented areas of the data distribution.

\subsubsection{Information density sampling}
For information density sampling, we aim to select samples that are both uncertain and located in densely populated regions of the feature space. We use the same logistic regression model as in entropy sampling with diversity to predict the probabilities of all the unlabeled samples, and compute the entropy of each sample to quantify its uncertainty. To incorporate information density, we also compute the pairwise distances between each unlabeled sample and all other unlabeled samples, assigning a density score based on how densely a sample is surrounded by other samples. The final score for each sample is determined by multiplying the entropy score and the density score, ensuring that the selected samples are informative and representative of dense regions in the feature space. 


\subsection{GPT as Oracle}
\label{sec:III.B}
GPT is used as an oracle in the active learning loop to provide labels for the queried instances. By leveraging GPT's extensive pre-trained knowledge, the need for human annotators is eliminated. To achieve this, prompt engineering is utilized to harness the question-answering capabilities of the GPT-4o model to issue labels. To ensure applicability across different tasks, we designed a structured prompt, as shown in Table \ref{TabPrompt1}. The prompt consists of several components: "A" specifies the classification task, "B" indicates whether it is a binary or multi-class classification, and "C" defines the classification standard. Finally, "query" contains the natural language text of instances selected by the active learning strategy. In conclusion, this design not only reduces reliance on human experts but also mitigates the limitations of using GPT directly for large-scale classification, such as high computation costs and long processing times.


\begin{table}
\centering
\caption{GPT Prompt Structure}
\label{TabPrompt1}
\resizebox{\columnwidth}{!}{%
\begin{tabular}{l} 
\toprule
\textbf{Structured Prompt}                                                                                                                                                                                                    \\ 
\hline
\begin{tabular}[c]{@{}l@{}}\{"role": "system", "content": f"You are an expert in \{A\}."\},\\ \{"role": "user", "content": f"Now you have a \{B\}. \\Please \{C\}:'\{query\}'. Please only return the label."\}\end{tabular}  \\
\bottomrule
\end{tabular}%
}
\end{table}
\vspace{1em}
\section{Evaluation}

\subsection{Research Questions}

To evaluate our approach, we focus on the following research questions:

\begin{itemize}
    \item RQ1: How does our approach perform across different tasks without using any human-labeled samples? \\
    
    \item RQ2: What is the impact of using GPT as an oracle in active learning, compared to traditional human expert labeling? \\
    
    \item RQ3: What are the advantages of our approach compared to directly using GPT for classification tasks? \\
    
    \item RQ4: What is the contribution of the active learning component in our approach? \\

\end{itemize}

\subsection{Data Preparation}
\label{sec:IV.B}
To evaluate our approach, we prepared three widely used datasets for three different text classification tasks: the IMDB movie reviews dataset for sentiment classification, the AGnews dataset for news categorization, and the Jigsaw Toxic Comment Classification dataset for toxic comment classification task.

Firstly, we preprocessed all three datasets to standardize the labels. In the IMDB dataset, "1" was used to represent positive sentiment, and "0" for negative sentiment. In the AGnews dataset, labels were assigned as follows: "0" for World, "1" for Sports, "2" for Business, and "3" for Sci/Tech. For the Jigsaw Toxic Comment Classification dataset, "0" was used for non-toxic comments and "1" for toxic comments.

\begin{figure}[]
  \centering
  \includegraphics[width=\linewidth]{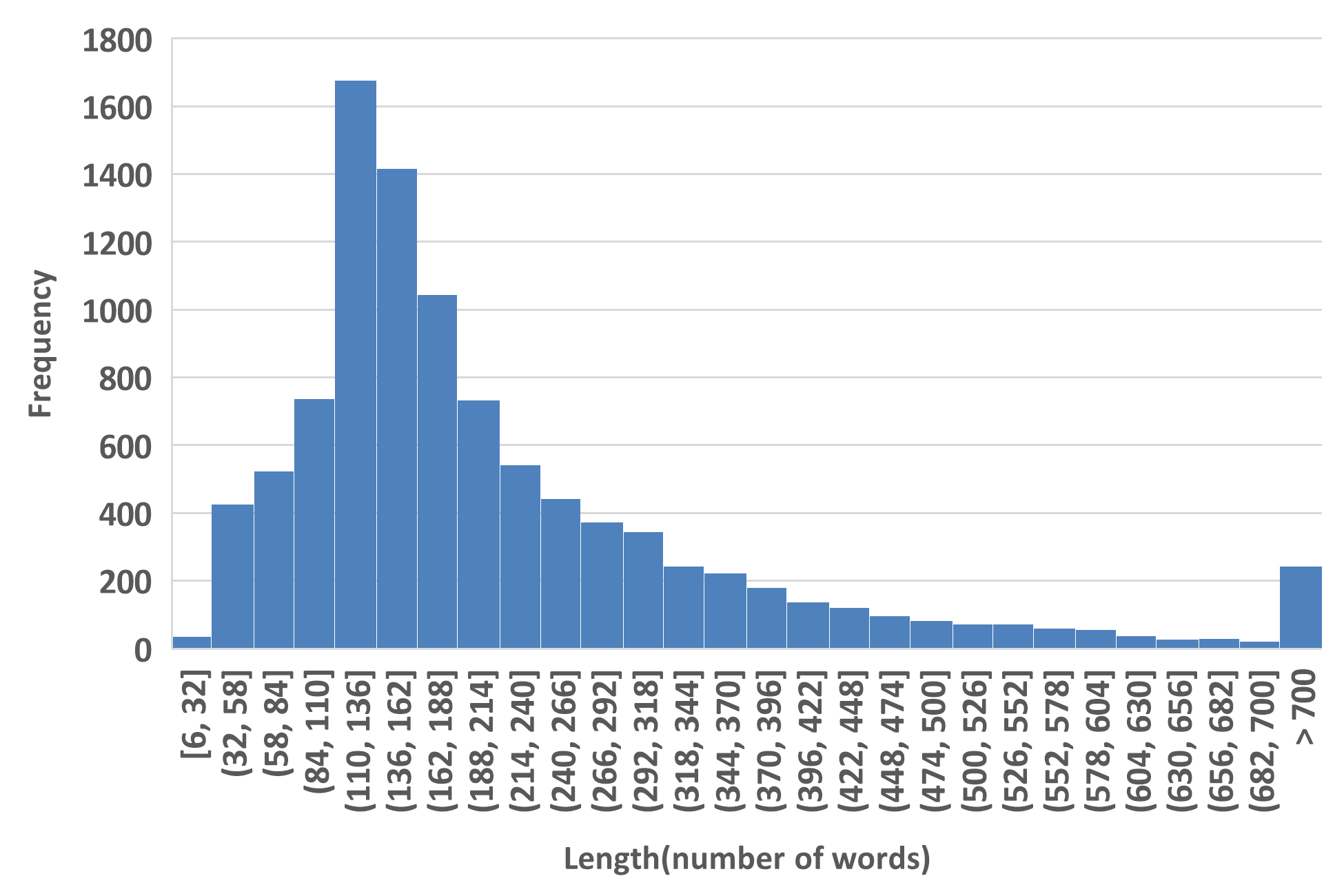}
  \caption{Length Distribution of IMDB Dataset}
  \label{FigIMDB}
\end{figure}

\begin{figure}[]
  \centering
  \includegraphics[width=\linewidth]{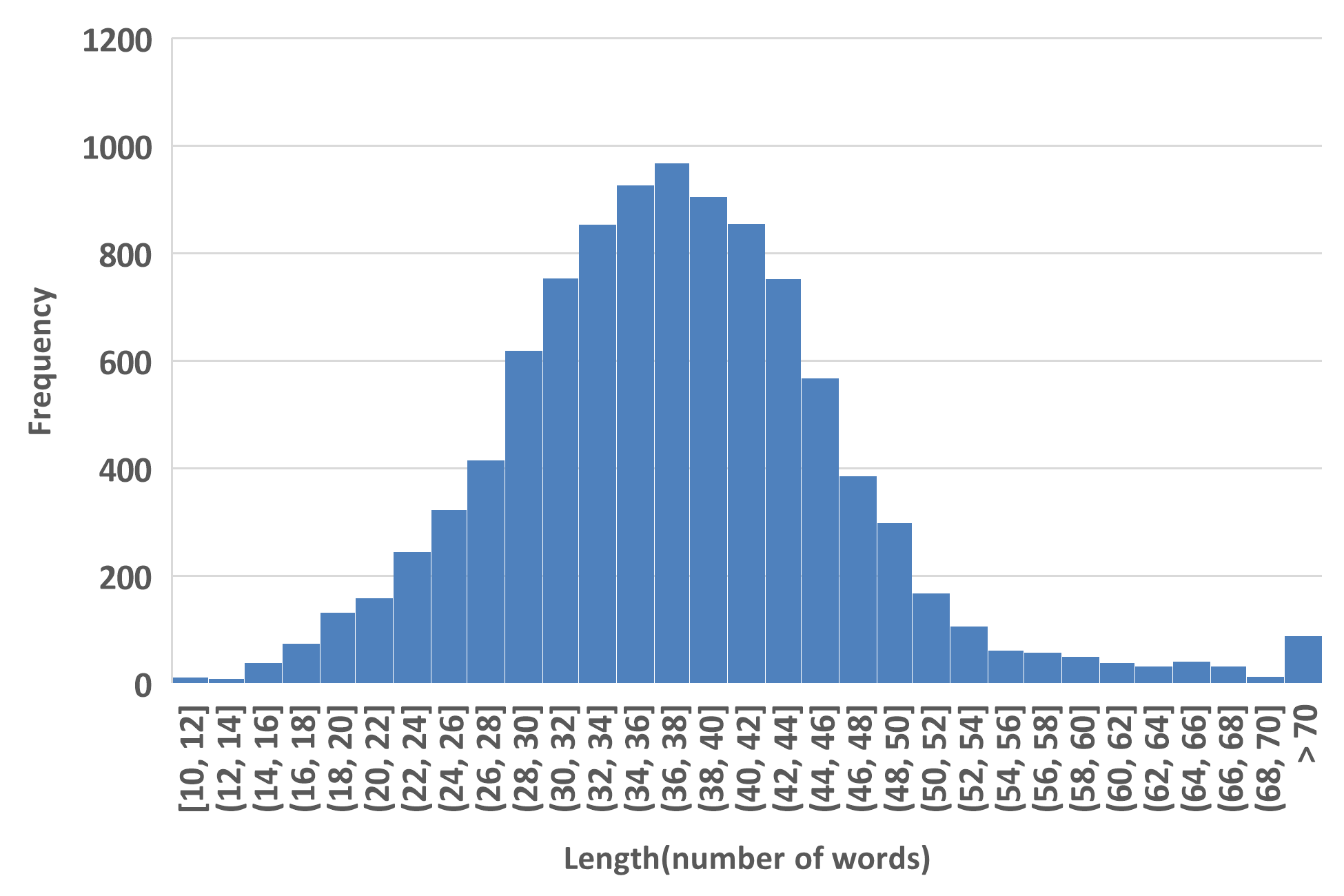}
  \caption{Length Distribution of AGnews Dataset}
  \label{FigAGnews}
\end{figure}

\begin{figure}[]
  \centering
  \includegraphics[width=\linewidth]{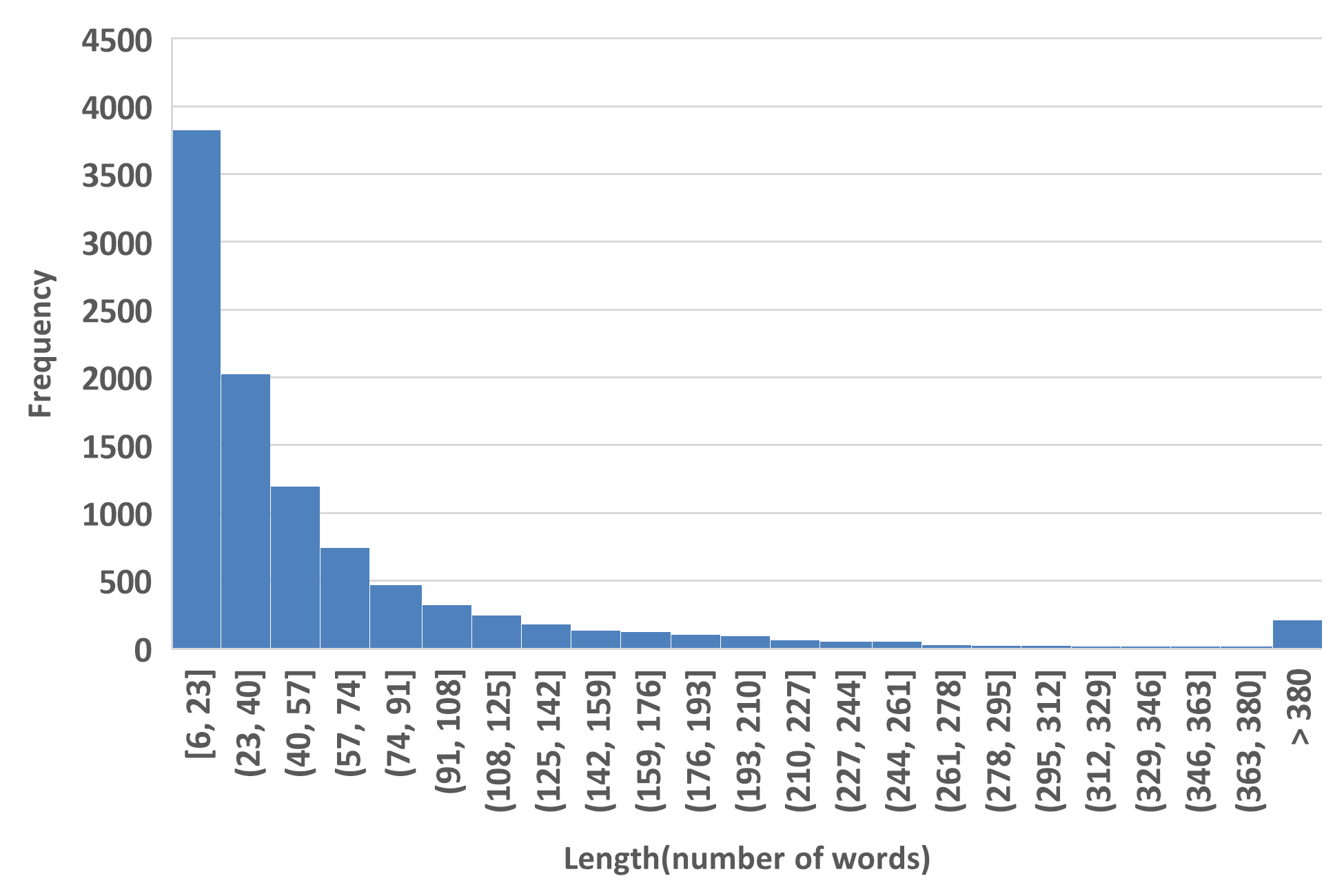}
  \caption{Length Distribution of Jigsaw Toxic Comment
Classification Dataset}
  \label{FigJig}
\end{figure}

\begin{figure}[]
  \centering
  \includegraphics[width=\linewidth]{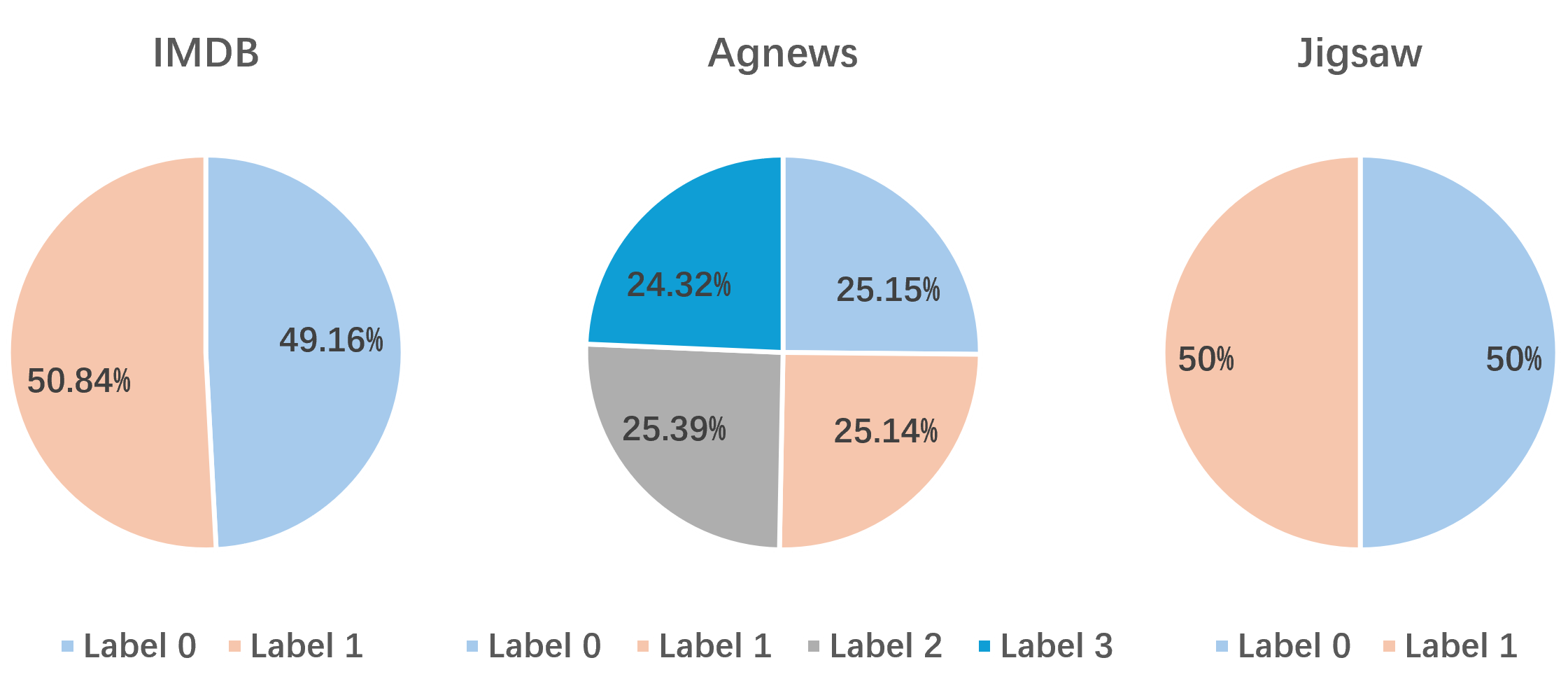}
  \caption{Label Distributions}
  \label{DataLabel}
\end{figure}

After label standardization, we randomly selected 10,000 instances from both the IMDB and AGnews datasets for our experiments.
As for the Jigsaw Toxic Comment Classification dataset, since it had issues such as data with very few words and significant class imbalance, we further processed the dataset by removing extremely short instances—specifically, those
with five words or fewer. After this filtering, we selected 5,000 instances each from toxic and non-toxic comments to create the final experimental dataset. The text length distributions of the three datasets are shown in Figure \ref{FigIMDB}, Figure \ref{FigAGnews}, and Figure \ref{FigJig}. The label distributions are summarized in Figure \ref{DataLabel}.

\subsection{Experimental Setup}
The implementation details of our approach have been discussed in Section \ref{sec:III}. This section focuses on the evaluation metrics, experimental environment, and structured prompt settings for the experiments based on the research questions.

\subsubsection{Evaluation Metrics}

We used accuracy as the primary evaluation metric, along with F1 score and recall to assess the performance of our experiments. For the multi-class task, we employed macro F1 and macro recall. The calculation of accuracy is given in Equation (\ref{eqAcc}). The F1 score for binary classification is calculated as shown in Equation (\ref{eqBF1}), and for multi-class classification, macro F1 is given by Equation (\ref{eqMF1}). Similarly, recall for binary classification is computed as in Equation (\ref{eqBrecall}), and macro recall for multi-class classification is calculated as in Equation (\ref{eqMrecall}).

\begin{equation}
\label{eqAcc}
\text{Accuracy} = \frac{\text{Number of Correct Predictions}}{\text{Total Number of Predictions}}
\end{equation}

\begin{equation}
\label{eqBF1}
\text{F1 (Binary)} = \frac{2 \times TP}{2 \times TP + FP + FN}
\end{equation}

\begin{equation}
\label{eqMF1}
\text{Macro F1} = \frac{1}{C} \sum_{i=1}^{C} \frac{2 \times TP_i}{2 \times TP_i + FP_i + FN_i}
\end{equation}

\begin{equation}
\label{eqBrecall}
\text{Recall (Binary)} = \frac{TP}{TP + FN}
\end{equation}

\begin{equation}
\label{eqMrecall}
\text{Macro Recall} = \frac{1}{C} \sum_{i=1}^{C} \frac{TP_i}{TP_i + FN_i}
\end{equation}
where \( TP \) (true positive) is the number of correctly predicted positive instances, \( FP \) (false positive) is the number of incorrectly predicted positive instances, and \( FN \) (false negative) is the number of missed positive instances. In the case of multi-class classification, \( C \) represents the total number of classes, and \( TP_i \), \( FP_i \), and \( FN_i \) correspond to the counts for class \( i \).

\subsubsection{Experimental Environment}

All experiments were conducted in the Google Colab environment, utilizing an NVIDIA A100-SXM4-40GB and an Intel(R) Xeon(R) CPU @ 2.20GHz.

\subsubsection{Structured Prompt Settings}
\label{IV.C.3)}
Based on the three datasets prepared for the classification tasks described in Section \ref{sec:IV.B}, we populated the structured prompt components A, B, and C, as shown in Table \ref{TabPrompt1} of Section \ref{sec:III.B}, for each task as follows:

\begin{itemize}
    \item \textbf{IMDB Sentiment Classification Task:} 
    \begin{itemize}
        \item A = "user reviews sentiment classification"
        \item B = "binary sentiment classification task"
        \item C = "classify the following user review into positive sentiment or negative sentiment, use 1 for positive and 0 for negative"
    \end{itemize}
    
    \item \textbf{AGnews News Classification Task:} 
    \begin{itemize}
        \item A = "news article classification"
        \item B = "four-class news topic classification task"
        \item C = "classify the following news article into one of the following categories: 0 for World, 1 for Sports, 2 for Business, or 3 for Sci/Tech"
    \end{itemize}
    
    \item \textbf{Jigsaw Toxic Comment Classification Task:} 
    \begin{itemize}
        \item A = "toxic comment classification"
        \item B = "binary classification task"
        \item C = "classify the following comment into toxic or non-toxic, use 1 for toxic and 0 for non-toxic"
    \end{itemize}
\end{itemize}

\vspace{2em}

\subsection{Evaluation for RQ1}

To address RQ1, we used all three datasets and evaluated our approach primarily using accuracy, with F1 score and recall as supplementary metrics. To ensure the robustness and reliability of the experimental results, we performed 5-fold cross-validation, with each fold using 80\% of the dataset for training and 20\% for testing. 

In each fold of the cross-validation, we initially randomly selected 50 instances (representing 0.625\% of the training set) from the training set as the seed set for starting active learning, and used structured prompts to request labels from GPT. We then began the active learning loop, setting the batch size to 5 and the number of iterations to 100. This means that each active learning loop consisted of 100 iterations, with each of the four query strategies independently selecting the top 5 most uncertain instances in each iteration, and requesting labels from GPT using our structured prompts. The obtained labels were then used to update each independent model.

\begin{figure}[H]
  \centering
  \includegraphics[width=\linewidth]{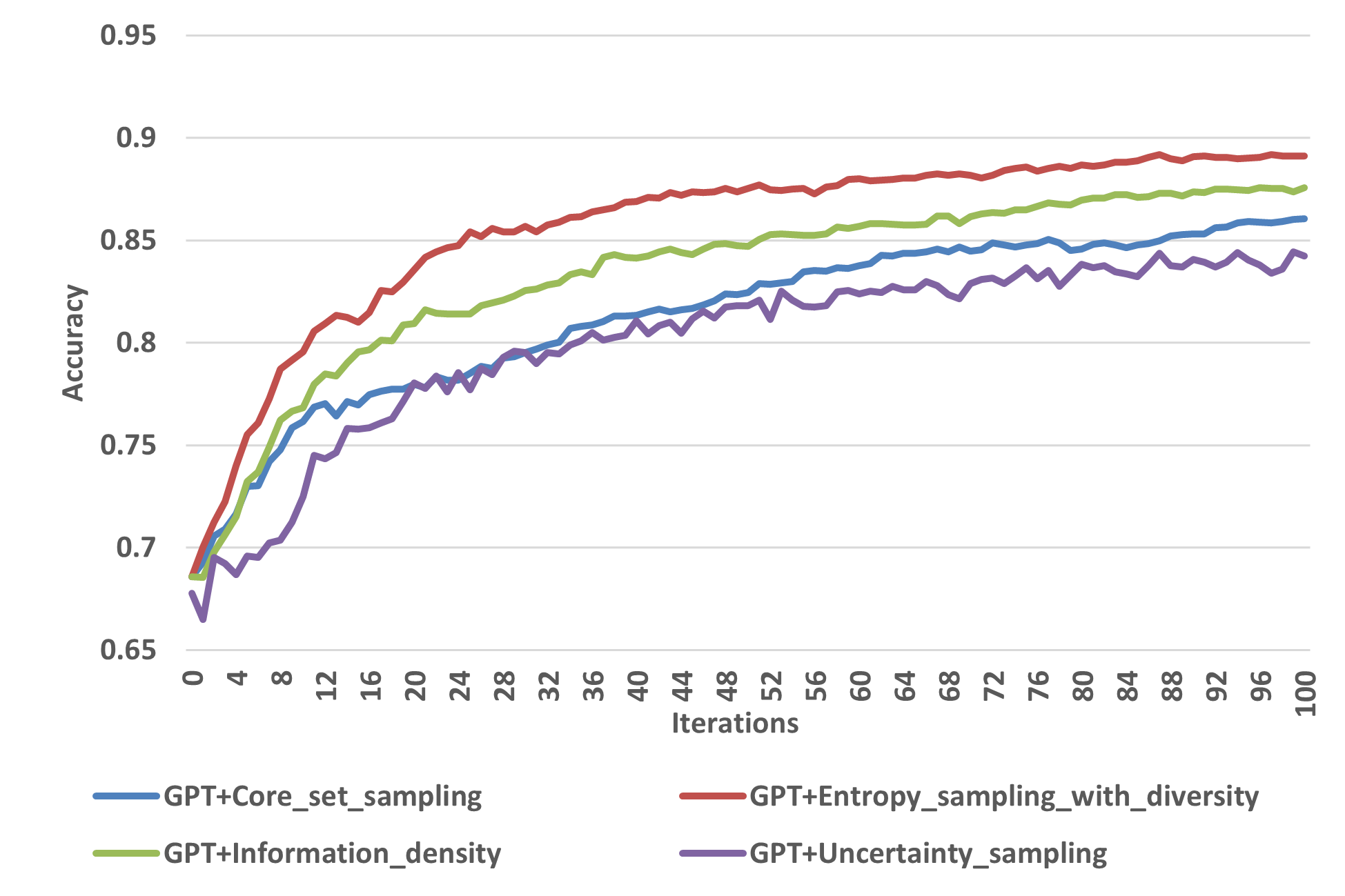}
  \caption{RQ1 IMDB Sentiment Classification Learning Curve}
  \label{FigRQ1IMDB}
\end{figure}

\begin{figure}[H]
  \centering
  \includegraphics[width=\linewidth]{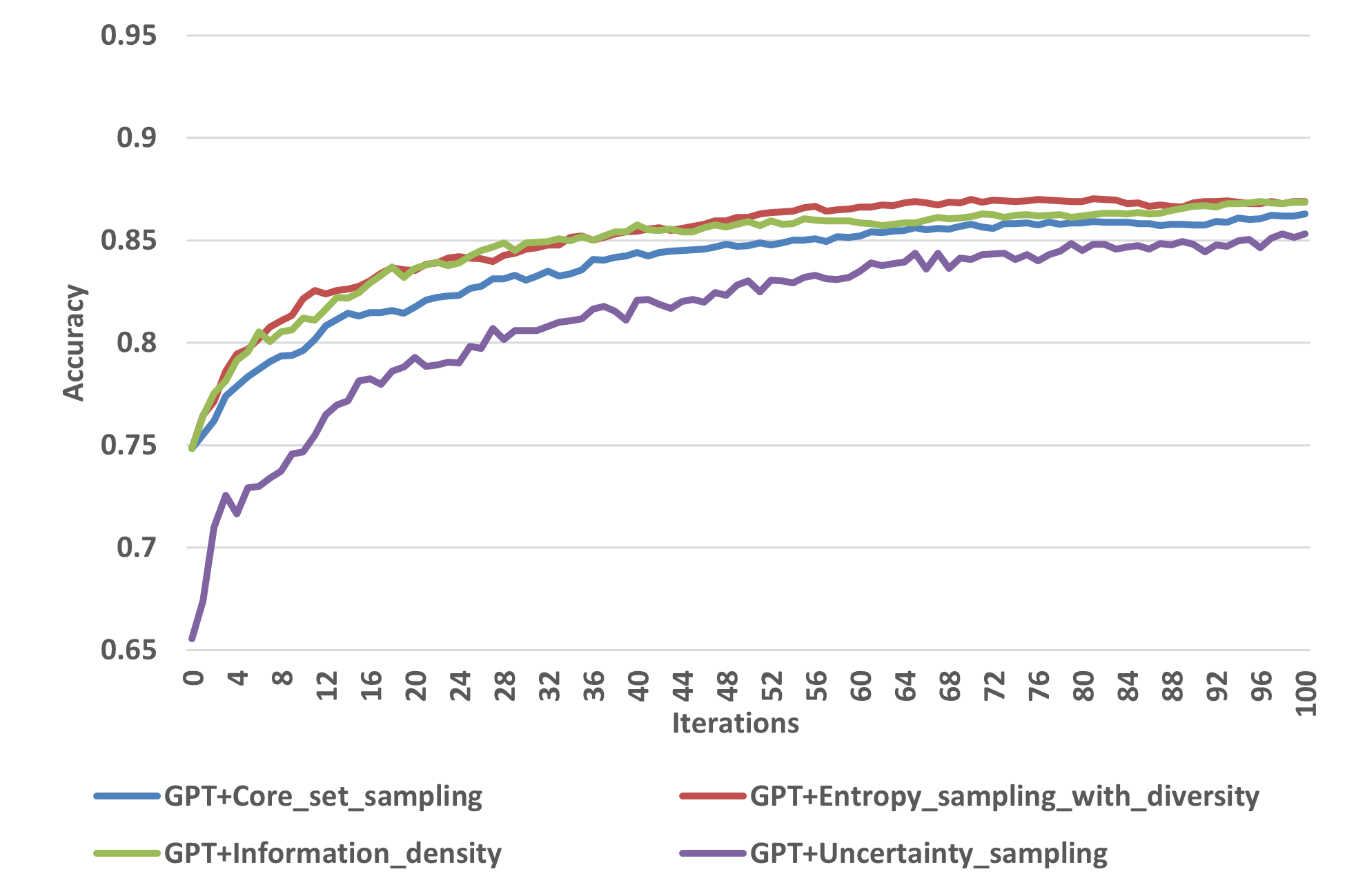}
  \caption{RQ1 AGnews News Classification Learning Curve}
  \label{FigRQ1AGnews}
\end{figure}

\begin{figure}[H]
  \centering
  \includegraphics[width=\linewidth]{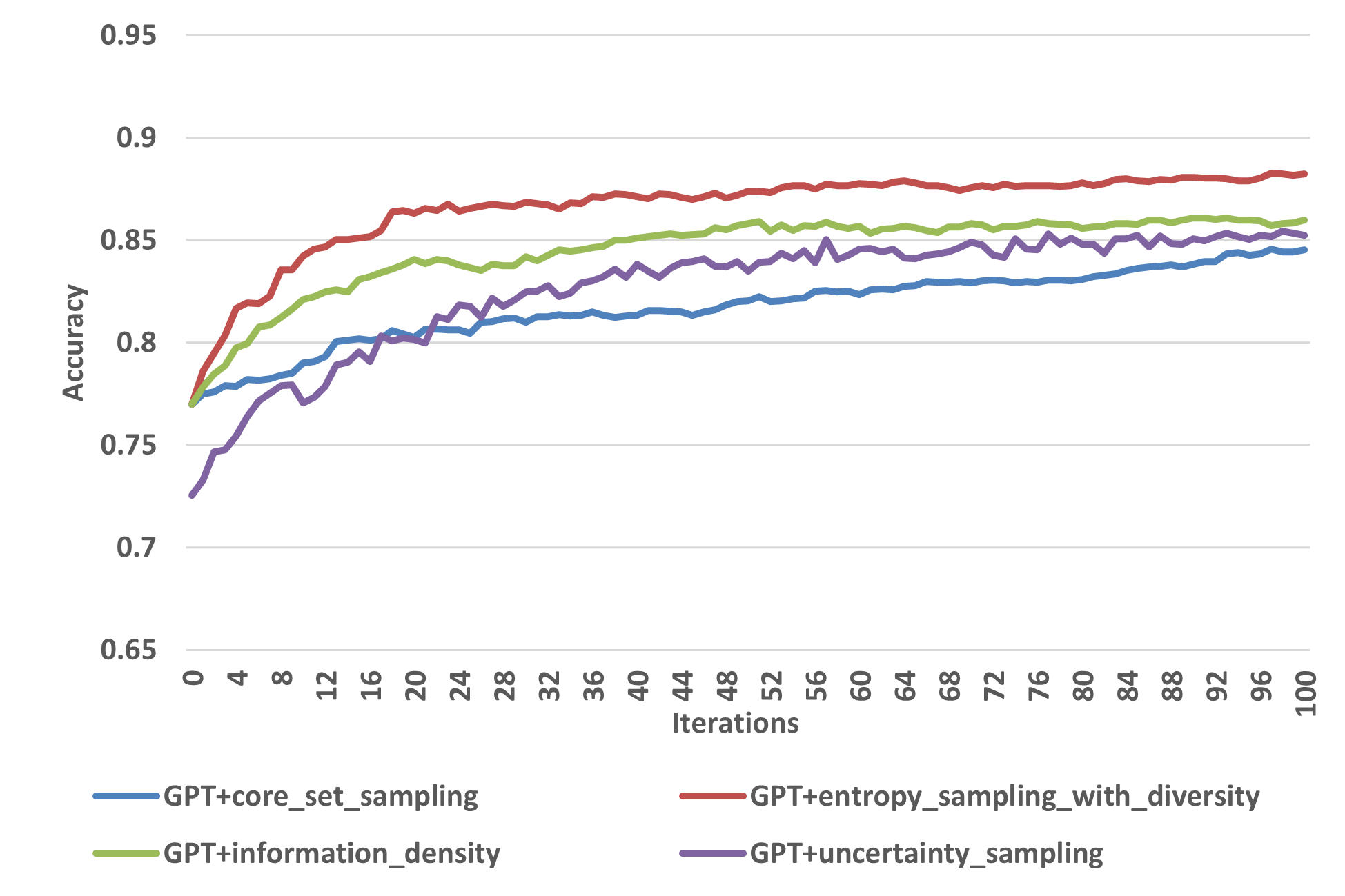}
  \caption{RQ1 Jigsaw Toxic Comment Classification Learning Curve}
  \label{FigRQ1Jig}
\end{figure}

Figures \ref{FigRQ1IMDB}, \ref{FigRQ1AGnews}, and \ref{FigRQ1Jig} respectively show the learning curve of our approach for the IMDB sentiment, AGnews, and Jigsaw toxic comment classification tasks.
The x-axis represents the number of iterations, and the y-axis represents the primary evaluation metric, accuracy, which is the average accuracy obtained from 5-fold cross-validation. The four lines represent the four query strategies combined with GPT. 

From Figure \ref{FigRQ1IMDB}, we observed that for the IMDB sentiment classification task, the combination of GPT with entropy sampling with diversity performed the best. Since active learning involves a trade-off between performance and labeling cost, we chose the 25th iteration—when the performance improvement began to plateau—as the final result for the sentiment classification task on the IMDB dataset, as recorded in Table \ref{TabRQ1IMDB}.

From Figure \ref{FigRQ1AGnews}, we observed that for the AGnews news classification task, both information density and entropy sampling with diversity demonstrated similar strong performance. We chose the data from the 30th iteration, where the performance growth of both methods began to plateau, and selected the higher performance (with information density showing a slight advantage at that point) as the final performance of our approach for the news classification task. The detailed results are recorded in Table \ref{TabRQ1AGnews}.

From Figure \ref{FigRQ1Jig}, we observed that for the Jigsaw toxic comment classification task, entropy sampling with diversity demonstrated an advantage. Similarly, since active learning involves a trade-off between labeling cost and performance, we selected the result from the 19th iteration, where the performance growth began to plateau, as the final toxic comment classification performance of our approach. The detailed results are presented in Table \ref{TabRQ1Jig}.

 In summary, our approach achieved notable classification performance without any human-labeled data. Specifically, on the IMDB sentiment classification task, our method obtained an accuracy of 85.42\% by using GPT for labels on 175 instances. For the AGnews news classification task, an accuracy of 84.88\% was achieved by using 200 labeled instances from GPT. Lastly, for the Jigsaw toxic comment classification task, an accuracy of 86.44\% was obtained by using 145 labeled instances from GPT. 
These results demonstrate that our approach effectively achieves high classification performance across different tasks without relying on human annotations, proving both its effectiveness and its cross-task generalization capability.


\begin{table}[H]
\centering
\caption{RQ1 IMDB Sentiment Classification Result}
\label{TabRQ1IMDB}
\begin{tblr}{
  hline{1,8} = {-}{0.08em},
  hline{2} = {-}{},
}
\textbf{Fold}    & \textbf{Accuracy} & \textbf{F1}     & \textbf{Recall} \\
1                & 0.8555            & 0.8555          & 0.8557          \\
2                & 0.8685            & 0.8680          & 0.8683          \\
3                & 0.8335            & 0.8335          & 0.8345          \\
4                & 0.8635            & 0.8635          & 0.8638          \\
5                & 0.8500            & 0.8497          & 0.8509          \\
\textbf{Average} & \textbf{0.8542}   & \textbf{0.8540} & \textbf{0.8546} 
\end{tblr}
\end{table}

\begin{table}[H]
\centering
\caption{RQ1 AGnews News Classification Result}
\label{TabRQ1AGnews}
\begin{tblr}{
  hline{1,8} = {-}{0.08em},
  hline{2} = {-}{},
}
\textbf{Fold}    & \textbf{Accuracy} & \textbf{F1}     & \textbf{Recall} \\
1                & 0.8345            & 0.8347          & 0.8351          \\
2                & 0.8285            & 0.8252          & 0.8258          \\
3                & 0.8500            & 0.8486          & 0.8495          \\
4                & 0.8695            & 0.8680          & 0.8684          \\
5                & 0.8615            & 0.8610          & 0.8604          \\
\textbf{Average} & \textbf{0.8488}   & \textbf{0.8475} & \textbf{0.8478} 
\end{tblr}
\end{table}

\begin{table}[H]
\centering
\caption{RQ1 Jigsaw Toxic Comment Classification Result}
\label{TabRQ1Jig}
\begin{tblr}{
  hline{1,8} = {-}{0.08em},
  hline{2} = {-}{},
}
\textbf{Fold}    & \textbf{Accuracy} & \textbf{F1}     & \textbf{Recall} \\
1                & 0.8450            & 0.8442          & 0.8449          \\
2                & 0.8665            & 0.8665          & 0.8666          \\
3                & 0.8705            & 0.8704          & 0.8704          \\
4                & 0.8725            & 0.8721          & 0.8721          \\
5                & 0.8675            & 0.8674          & 0.8679          \\
\textbf{Average} & \textbf{0.8644}   & \textbf{0.8641} & \textbf{0.8644} 
\end{tblr}
\end{table}

\subsection{Evaluation for RQ2}

\subsubsection{Setup}
To evaluate the impact of GPT as an oracle in active learning, we compared the following three scenarios:

\begin{enumerate}
    \item \textbf{Full GPT Labeling}: GPT is used for labeling both the seed set and all queries.
    \item \textbf{Hybrid Labeling}: Human annotators provide labels for the seed set, while GPT labels all queries.
    \item \textbf{Full Human Labeling}: Human annotators label both the seed set and all queries.
\end{enumerate}

For scenarios requiring human-labeled samples, we directly used the labels from the original dataset. To ensure fairness in the comparative experiments, we used the same 5-fold splits and conducted experiments on the three classification tasks using the same experimental setup as in RQ1.

\begin{figure}[H]
  \centering
  \includegraphics[width=\linewidth]{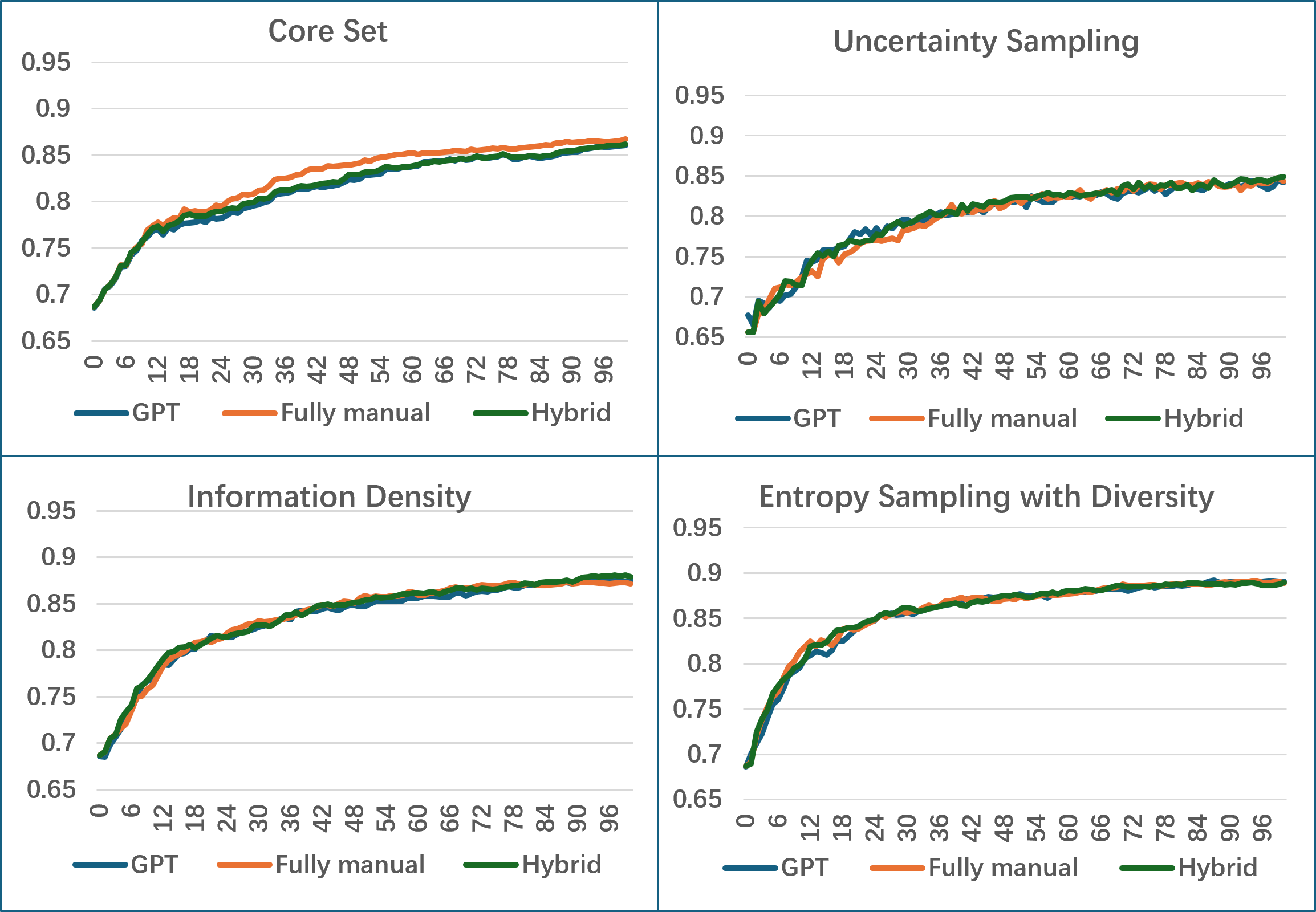}
  \caption{RQ2 IMDB Sentiment Classification Learning Curves}
  \label{FigRQ2IMDB}
\end{figure}
\vspace{1em}
\begin{figure}[H]
  \centering
  \includegraphics[width=\linewidth]{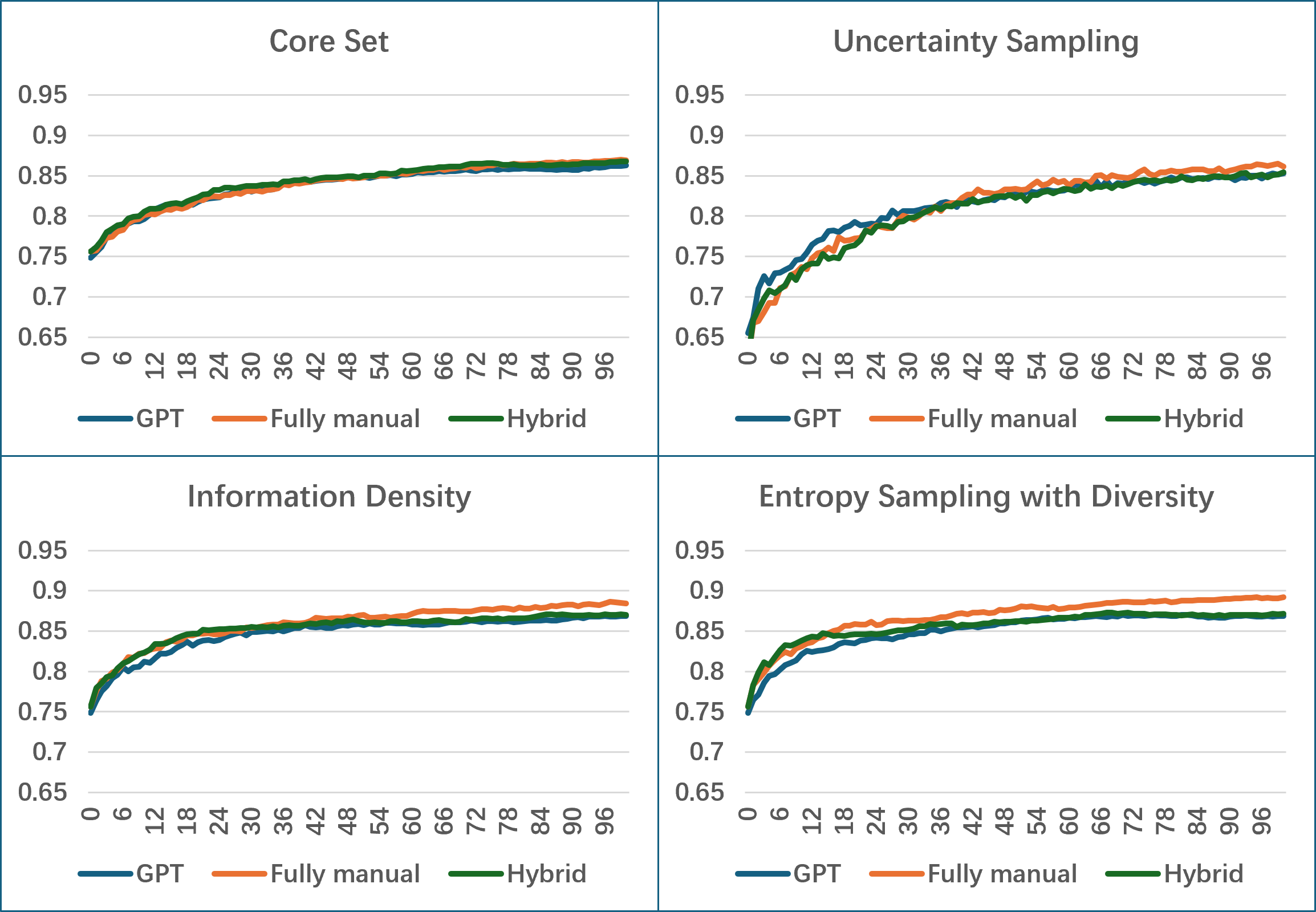}
  \caption{RQ2 AGnews News Classification Learning Curves}
  \label{FigRQ2AGnews}
\end{figure}
\vspace{1em}
\begin{figure}[H]
  \centering
  \includegraphics[width=\linewidth]{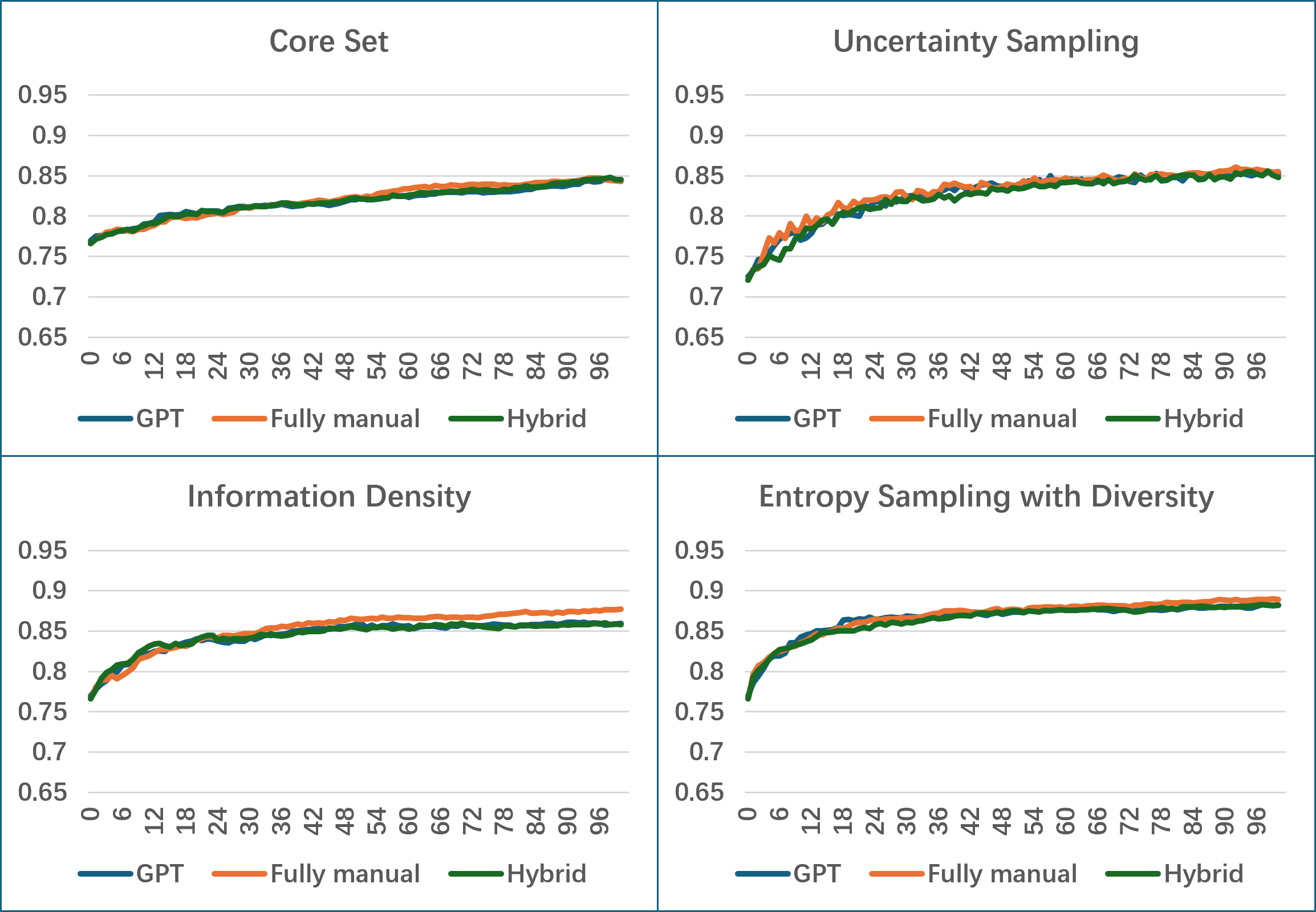}
  \caption{RQ2 Jigsaw Toxic Comment Classification Learning Curves}
  \label{FigRQ2Jig}
\end{figure}


\subsubsection{Results}

Figures \ref{FigRQ2IMDB}, \ref{FigRQ2AGnews}, and \ref{FigRQ2Jig} respectively show the comparative learning curves for the four query strategies on the IMDB sentiment, AGnews, and Jigsaw toxic comment classification task. Similar to RQ1, the y-axis represents the average accuracy from the 5-fold cross-validation, while the x-axis represents the number of iterations. Each figure contains four subplots, each representing a corresponding query strategy. In each subplot, three lines are shown, representing the cases of full GPT labeling, hybrid labeling, and full human labeling.

To provide a more detailed understanding of the comparative learning curves, we first calculated the standard deviation of the accuracy obtained from full GPT labeling, full human labeling, and hybrid labeling at each iteration. Then, we averaged these standard deviations across all iterations within each query strategy to obtain the average standard deviation for each query strategy in each of the three classification tasks. Detailed standard deviation results are listed in Table \ref{TabRQ2STD}. For each query strategy on each task, the maximum cross iteration average standard deviation observed was 0.007961, and the overall average standard deviation across all tasks and query strategies was 0.004056.
The relatively small average standard deviation across all query strategies and tasks indicates that the accuracy differences among full human labeling, hybrid labeling, and full GPT labeling remain stable across iterations.
\begin{table}[H]
\centering
\caption{RQ2 Average Standard Deviation}
\label{TabRQ2STD}
\resizebox{\columnwidth}{!}{%
\begin{tabular}{|l|l|l|l|l|} 
\hline
                & \textbf{Core Set} & \begin{tabular}[c]{@{}l@{}}\textbf{Entropy\&}\\\textbf{Diversity}\end{tabular} & \textbf{Information} & \textbf{Uncertainty}  \\ 
\hline
\textbf{IMDB}   & 0.004939          & 0.002142                                                                       & 0.002463             & 0.003870              \\ 
\hline
\textbf{AGnews} & 0.002546          & 0.007961                                                                       & 0.005305             & 0.005883              \\ 
\hline
\textbf{Toxic}    & 0.002156          & 0.002818                                                                       & 0.004879             & 0.003715              \\
\hline
\end{tabular}%
}
\end{table}

Furthermore, we calculated the accuracy differences between GPT labeling and full human labeling, as well as between GPT labeling and partial human labeling, at each iteration. We then averaged these accuracy differences across all iterations within each query strategy to obtain the average accuracy difference for each query strategy in each of the three classification tasks. Detailed accuracy difference results are shown in Table \ref{RQ2AVE}, where negative values indicate that the accuracy of the active learning loop using full GPT labeling was lower, whereas positive values indicate higher accuracy. 
The worst case for full GPT labeling against full human labeling was -1.7321\%, while the average across all tasks and strategies was -0.5168\%. Similarly, the worst case of using full GPT labeling against hybrid labeling was -0.5985\%, and the overall average was -0.1264\%.
These results demonstrate that, despite minor fluctuations, the overall accuracy disadvantage of full GPT labeling is also small, reinforcing the observation that GPT labeling performs comparably to both human labeling and hybrid labeling across the evaluated tasks.

\begin{table}[H]
\centering
\caption{RQ2 Comparison of Average Accuracy Differences}
\label{RQ2AVE}
\resizebox{\columnwidth}{!}{%
\begin{tblr}{
  cell{1}{1} = {c=2}{},
  cell{2}{1} = {r=3}{},
  cell{5}{1} = {r=3}{},
  vlines,
  hline{1-2,5,8} = {-}{},
  hline{3-4,6-7} = {2-6}{},
}
                                                    &        & {\textbf{Core }\\\textbf{Set}} & {\textbf{Entropy}\\\textbf{\&Diversity}} & {\textbf{Info\textbf{r}-}\\\textbf{mation}} & {\textbf{Uncer-}\\\textbf{tainty}} \\
{\textbf{GPT vs }\\\textbf{Fully}\\\textbf{Manual}} & IMDB   & -0.011189                      & -0.002010                                & -0.002209                                   & ~0.001413                          \\
                                                    & AGnews & -0.002059                      & -0.017321                                & -0.011646                                   & ~0.000159                          \\
                                                    & Toxic    & -0.002556                      & -0.003018                                & -0.007780                                   & -0.003796                          \\
{\textbf{GPT vs}\\\textbf{Semi-}\\\textbf{manual}}  & IMDB   & -0.002429                      & -0.001682                                & -0.003287                                   & -0.001831                          \\
                                                    & AGnews & -0.004733                      & -0.005985                                & -0.005905                                   & ~0.006666                          \\
                                                    & Toxic    & -0.000639                      & ~0.001758                                & -0.000205                                   & ~0.003101                          
\end{tblr}%
}
\end{table}

\subsubsection{Conclusion}

Based on these results, we conclude that there is no substantial difference in the overall performance of active learning when using full human labeling, hybrid labeling, or fully GPT-labeled data.

It is also important to note that in simulating human labeling, we used the ground truth labels from the original datasets. Since these datasets are well-known and have been widely used in academic research, we assume that the labeling quality is high. In other labeling scenarios, the quality of human labeling might not reach the same level of rigor.

In conclusion, given the comparable performance between using GPT labeling and human labeling for active learning, GPT labeling offers a viable and cost-effective alternative for large-scale annotation tasks, particularly when human resources are limited.

\subsection{Evaluation for RQ3}

To address RQ3, we evaluated the efficiency and cost-effectiveness of our proposed approach by comparing it with a baseline where GPT is directly used as a classifier without any active learning or model training. Specifically, we employed the structured prompt described in Section \ref{IV.C.3)} to obtain predictions for the test set.

To ensure fairness in the comparative experiments, we used the same 5-fold cross-validation split as in RQ1. For each dataset, we recorded the following metrics:
\begin{itemize}
    \item \textbf{Classification Accuracy}: The average accuracy across five folds.
    \item \textbf{Time Expenditure}: The average classification time across five folds.
    \item \textbf{Monetary Cost}: The average cost of GPT API usage across five folds, calculated based on the number of tokens processed.
\end{itemize}

Table \ref{tab:RQ3_results} summarizes the 5-fold average classification accuracy, time expenditure, and cost for both direct GPT usage and our proposed approach across the three datasets.

\begin{table}[H]
\centering
\caption{RQ3 Comparison Between Our Approach and Direct GPT Usage}
\label{tab:RQ3_results}
\resizebox{\columnwidth}{!}{%
\begin{tblr}{
  cell{1}{1} = {r=2}{},
  cell{1}{2} = {c=3}{c},
  cell{1}{5} = {c=3}{c},
  vlines,
  hline{1,3-6} = {-}{},
  hline{2} = {2-7}{},
}
                & \textbf{Direct GPT Usage } &               &               & \textbf{Our Approach } &               &               \\
                & Accuracy~                  & {Time\\(min)} & {Cost\\(USD)} & Accuracy~              & {Time\\(min)} & {Cost\\(USD)} \\
\textbf{IMDB}   & 0.9463                     & 17.81         & 3.23          & 0.8542                 & 1.29        & 0.20          \\
\textbf{AGnews} & 0.8866                     & 15.14         & 1.17          & 0.8488                 & 0.89          & 0.09          \\
\textbf{Jigsaw} & 0.9178                     & 16.20         & 1.58          & 0.8644                 & 0.80          & 0.06          
\end{tblr}
}
\end{table}

\vspace{3em}

Based on the experimental results, our approach required only 6\% of the time and monetary cost compared to directly using GPT for classification, while still achieving over 93\% of its classification performance. These findings demonstrate that our approach is not only significantly more cost-effective but also highly efficient, making it particularly advantageous for cost-sensitive applications.

\subsection{Evaluation for RQ4}

To answer RQ4, we disabled the query strategy components and instead randomly selected the same number of instances from the training set as the combined total of the seed and query instances used in each of the final performance settings of RQ1. We used the sampled instances to train the default logistic regression classifier from the active learning module. To ensure the rigor of the comparative experiment, we conducted the same 5-fold cross-validation, and for each fold, we repeated the random sampling five times, taking the average as the result for that fold. The detailed accuracy results are shown in Table \ref{TabRQ4Random}.

\begin{table}[h]
\centering
\caption{RQ4 Performance without Active Learning}
\label{TabRQ4Random}
\resizebox{\columnwidth}{!}{%
\begin{tabular}{|l|l|l|l|l|l|l|} 
\hline
                & \textbf{Fold1} & \textbf{Fold2} & \textbf{Fold3} & \textbf{Fold4} & \textbf{Fold5} & \textbf{Average}  \\ 
\hline
\textbf{IMDB}   & 0.7515         & 0.7495         & 0.8010         & 0.7880         & 0.7880         & 0.7756            \\ 
\hline
\textbf{AGnews} & 0.7840         & 0.7945         & 0.7995         & 0.8140         & 0.8225         & 0.8029            \\ 
\hline
\textbf{Toxic}  & 0.8000         & 0.8045         & 0.7880         & 0.8010         & 0.8075         & 0.8002            \\
\hline
\end{tabular}%
}
\end{table}

As observed, incorporating the active learning component led to an average accuracy improvement of 7.86\% in the IMDB sentiment classification task, 4.59\% in the AGnews news classification task, and 6.42\% in the Jigsaw toxic comment classification task. This demonstrates that active learning contributes to model performance by guiding the selection of informative instances. Therefore, we conclude that integrating active learning into our approach is beneficial, as it enhances classification accuracy with a limited number of labeled instances.

\section{Conclusion}

This study proposes a cost-efficient text classification approach that integrates large language models (LLMs) with active learning, enabling effective cross-task text classification without requiring manually labeled data. Experimental results confirm that our approach successfully eliminates the need for human annotations while ensuring high classification performance across diverse tasks. 


Furthermore, we systematically evaluated the role of GPT as an oracle in active learning and quantified the contribution of different query strategies within the overall framework. Moreover, our experiments demonstrate that compared to directly using generative AI models like GPT for text classification, our approach achieves comparable performance while significantly reducing the cost.

For future work, we believe that our approach has high scalability potential. We plan to further evaluate its effectiveness in a wider range of text classification tasks, such as assessing its performance in multilingual settings and validating its applicability across additional datasets.

\bibliographystyle{IEEEtran}
\bibliography{references}

\EOD

\end{document}